\documentclass[letterpaper, 10 pt, conference]{ieeeconf}  

\IEEEoverridecommandlockouts                              

\usepackage{algorithm}
\usepackage{amsmath}
\usepackage{cite}
\usepackage{algpseudocode}
\usepackage{makecell}
\usepackage{amssymb}
\usepackage{caption}
\usepackage{subcaption}
\usepackage{bm}
\usepackage{hyperref}
\hypersetup{
    colorlinks=true,
    linkcolor=blue,
    filecolor=magenta,      
    urlcolor=black,
    pdftitle={Overleaf Example},
    pdfpagemode=FullScreen,
    }
\usepackage{gensymb}
\usepackage{graphicx}
\graphicspath{{./images/}}

\title{\LARGE \bf
SpinDOE: A ball spin estimation method for table tennis robot
}

\author{Thomas Gossard$^{1}$, Jonas Tebbe$^{1}$, Andreas Ziegler$^{1}$, Andreas Zell$^{1}$
\thanks{$^{1}$The authors are with the Cognitive Systems Group, Dept. Informatics, University of Tuebingen. Corresponding author {\tt\small thomas.gossard@uni-tuebingen.de}}%
\thanks{This research was funded by Sony AI.}
}

\begin{document}

\maketitle
\thispagestyle{empty}
\pagestyle{empty}

\begin{abstract}

Spin plays a considerable role in table tennis, making a shot's trajectory harder to read and predict. However, the spin is challenging to measure because of the ball's high velocity and the magnitude of the spin values. Existing methods either require extremely high framerate cameras or are unreliable because they use the ball's logo, which may not always be visible. Because of this, many table tennis-playing robots ignore the spin, which severely limits their capabilities.
This paper proposes an easily implementable and reliable spin estimation method. We developed a dotted-ball orientation estimation (DOE) method, that can then be used to estimate the spin. The dots are first localized on the image using a CNN and then identified using geometric hashing. The spin is finally regressed from the estimated orientations. Using our algorithm, the ball's orientation can be estimated with a mean error of $2.4\degree$ and the spin estimation has an relative error lower than 1\%. Spins up to 175 rps are measurable with a camera of 350 fps in real time. Using our method, we generated a dataset of table tennis ball trajectories with position and spin, available on our project page.

Project page: \url{https://cogsys-tuebingen.github.io/spindoe/}

\end{abstract}

\section{INTRODUCTION}

In table tennis, spin estimation is primordial to play the ball back correctly and win the game. Ball trajectories can indeed be made difficult to predict with spin. The ball will accelerate after bouncing with a top spin, and its airborne trajectory will curve sideways if sidespin is applied. In order to develop a table tennis-playing robot, spin estimation is thus paramount, not only during the match but also to build an accurate model of table tennis dynamics (aerodynamics, table bounce, racket bounce).

Human players can use different cues to estimate the ball's spin. Some players can get an idea of the spin from the motion blur generated by the logo on the ball \cite{2018}. However, most players use the opponent's stroke motion and prior knowledge of the rubber used (sticky rubber, anti-topspin rubber, pimpled rubber) to estimate how much spin was applied on the ball. The spin can also be estimated from its airborne trajectory and bounce, though this method leaves less time for the player to react.

Because of the difficulty of spin estimation, many table tennis robots choose to either ignore the ball spin\cite{abeyruwan2022}\cite{zhang2012}\cite{buchler2020} or estimate it "implicitly" \cite{yang2021}. Only Tebbe et al. \cite{tebbe2020} explicitly estimate the ball's spin and take it into consideration when having the robot play.
 
Different methods have already been investigated for spin estimation. There are three main approaches to spin estimation, imitating human skills:  observing the stroke (racket/body pose), analyzing the ball's trajectory, and directly observing the ball. 

Strokes can be classified from body pose estimation \cite{sato}\cite{kulkarni2021}, racket IMU measurements \cite{blank2015a} \cite{blank2017a} or racket pose estimation \cite{gao2021}. The issue with these methods is that they only give the type of spin applied (topspin, backspin, left or right sidespin, no spin). The exact spin of the ball (spin vector $\bm{\omega}$) is still unknown. For more accurate spin estimation,  most research has focused on extracting the spin from the ball's trajectory or a sequence of images of the ball. There has also been a method that extracts the spin from the motion blur generated by the ball's logo \cite{2008a}. However, this last method requires the blur to be essentially generated by the spin and not the velocity. That is not the case for table tennis. Indeed, when shooting balls with a ball gun, we noticed motion blur appearing when increasing the ball's velocity for similar spin values.

Because measuring the spin directly with cameras requires a high frame rate, many researchers focused on estimating the spin from the trajectory deviation caused by the Magnus effect\cite{chen2010a} \cite{tebbe2020} \cite{su2013}. However, they rarely had access to ground truth spin values to calculate the Magnus effect coefficient. Moreover, the spin estimation highly depends on the accuracy of the recorded ball positions. Because Magnus effect-caused deviations are minimal for low-magnitude spins, this spin estimation method becomes highly sensitive to position measurement noise and bias for low spin values. 

\begin{figure}
    \centering
    \includegraphics[width=0.5\textwidth]{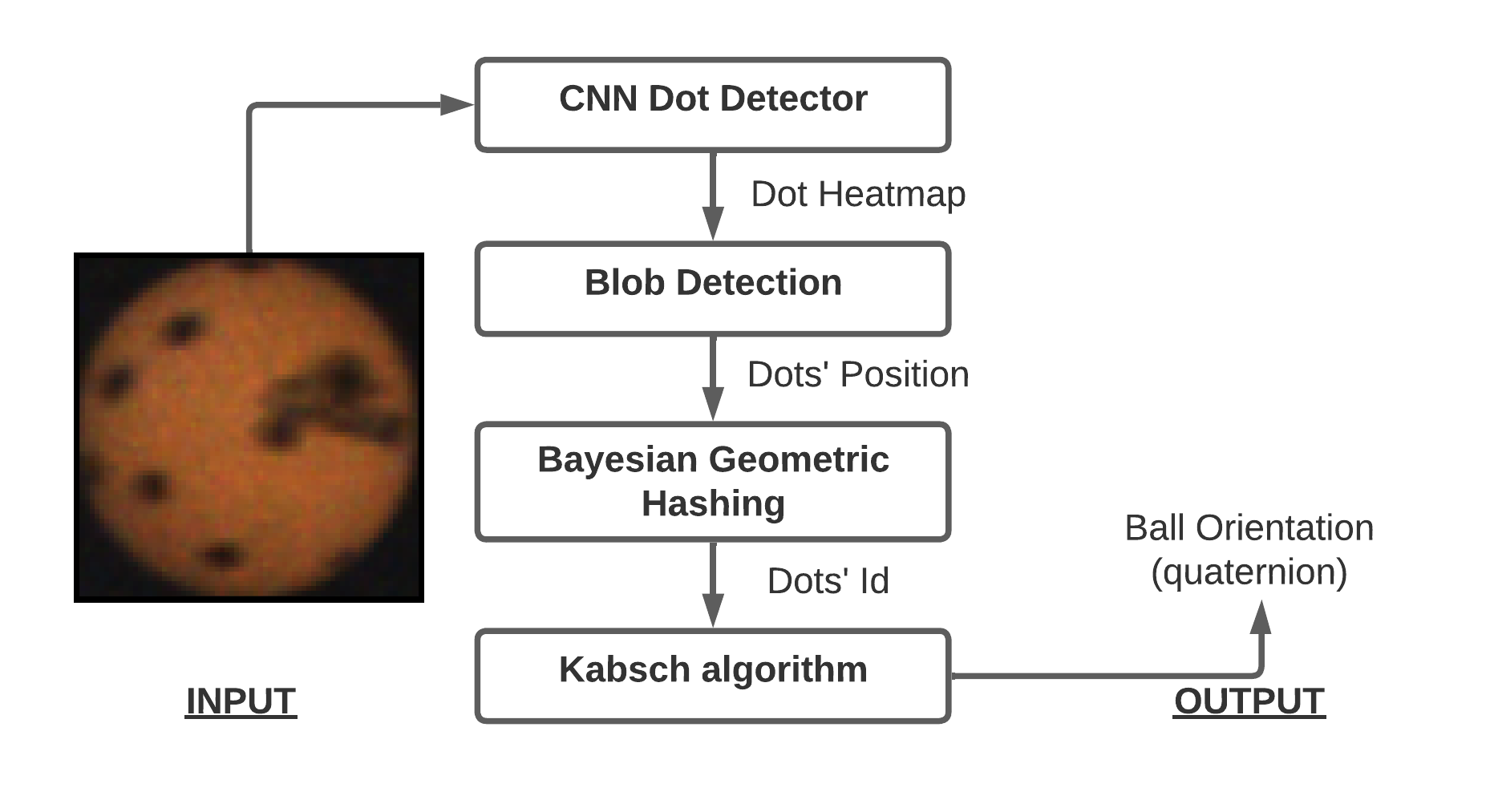}
    \caption{Orientation estimation pipeline (DOE)}
    \label{fig:orientation_pipeline}
\end{figure}

Ball observing-based methods are those that give the highest spin accuracy. They can be divided in 2 subcategories: logo-based \cite{zhang2015} \cite{tebbe2020} \cite{glover2014a} or pattern-based \cite{tamaki2004a} \cite{tamaki2012} \cite{furuno2009} \cite{szep2011}. Logo-based methods are highly interesting because they can be used for official table tennis matches. They use the logo to get an orientation of the ball, and they then fit a spin. However, because the logo is not always visible and often has some symmetry, spin estimation may often be impossible or very difficult (when the logo is only on the edge of the visible part of the ball or when the logo is close to the rotation axis). This can be compensated for by using 2 cameras to observe the ball from 2 different points of view \cite{zhang2015}, but it also introduces more complexity and cost to the perception system. The pattern-based approach is much more appropriate for research because of its higher reliability. Some approaches rely on registration \cite{tamaki2004a} \cite{tamaki2012} \cite{szep2011}. The problem with using registration is the necessity of a high enough resolution image to generate descriptors to identify the keypoints uniquely.

In this paper, we propose a new Dotted-ball Orientation Estimation method (DOE). DOE is then leveraged to estimate a table tennis ball's spin. We therefore named our method SpinDOE.

\begin{table*}
    \centering
    \caption{Summary of existing logo or patter-based spin estimation methods (papers are ordered in chronological order) }
    \label{fig:method_comparaison}
    \begin{tabular}{ | l  r | c | c | c | c | } 
     \hline
     \multicolumn{2}{|c|}{\textbf{Algorithm}} & \textbf{Method} & \textbf{Max spin (reported)[rps]} & \textbf{Camera fps} & \textbf{Relative error [\%]}\\ [0.5ex] 
     \hline\hline
     Tamaki et al & \cite{tamaki2004a} & Pattern (Registration) & ~90 & 500 &  NA \\ 
     \hline
     Theobalt et al & \cite{theobalt} & Pattern (Colored markers) & 27 & 80 & NA \\ 
     \hline
     Furuno et al & \cite{furuno2009} & Pattern (colored lines) & 23 & 1200  & NA \\ 
     \hline
     Boracchi et al & \cite{2008a} & Blur& 28 & NA & 3 \\ 
     \hline
     Szèp & \cite{szep2011} & Pattern (track corners) & 63 & 1000 &  12.5\\ 
     \hline
     Tamaki et al & \cite{tamaki2012} & Pattern  & 67 & 600 & NA\\ 
     \hline
     Glover et al & \cite{glover2014a} & Logo$^*$ & 50 & 200 & NA\\ 
     \hline
     Zhang et al & \cite{zhang2015} & Logo & 60 & NA & 0.07 \\ 
     \hline
     Tebbe et al & \cite{tebbe2020} & Logo & 75 & 380 & NA\\
     \hline
     \textbf{Our method} & & Pattern & 175 & 350 &  1\\ [1ex] 
     \hline
    \end{tabular}

\end{table*}

Our main \textbf{contributions} with SpinDOE are:
\begin{itemize}
    \item Having spin estimation work with a standard industrial camera.
    \item High reliability and accuracy for spins up to 175 rps 
    \item Making this method easy to reproduce with any kind of ball by providing the 3D printed stencil model and the source code.
    \item Proving the constant spin assumption while the ball is airborne to be correct.
    \item Generating the first published dataset of table tennis ball trajectories with accurate position and spin. 
\end{itemize}

In the rest of this paper, we will present how SpinDOE works. In section \ref{orientationestimation}, we first present our orientation estimation pipeline. Then in section \ref{spinestimation}, we show how we regress the spin from a sequence of orientations. In section \ref{experiments}, we expose our experimental results, such as an estimate of the spin dampening coefficient. The final section \ref{dataset} is dedicated to the creation of our table tennis ball trajectory dataset.

\section{ORIENTATION ESTIMATION} \label{orientationestimation}
We want to accurately estimate any ball's spin during an exchange without hindering the players. This means the camera has to be situated far from the table and requires a wide field of view. Because of these restrictions, captured images of the ball have a low resolution. Moreover, though the exposure time is reduced to the minimum, motion blur can still be observed for balls played at high speeds. This makes it impossible to use registration-based methods or to track multiple keypoints. Since we always want to be able to measure the ball's spin with only one camera, logo-based methods are also not an option. We thus decided to use a dot pattern to estimate the ball's orientation. A dot pattern has the advantage of not interfering with the ball detection pipeline. We can find the rotation between the reference 3D configuration of the dots and the measured 3D position of the dots. The difficulty of this approach comes from uniquely identifying the dots. Indeed, the observed dots need to be identified to be associated with their reference value. This is achieved using geometric hashing \cite{wolfson1997}. Instead of using the image features to identify them, like for registration, we use their spatial configuration. 

The DOE works as follows: a CNN first takes the ball's image and returns a heatmap of the dots' likely locations. The CNN dot detection is based on CenterNet \cite{duan2019a}. The dot location are then extracted from the heatmap via blob detection. In order to match the observed dots with the dots from the reference pattern, geometric hashing is used. We can finally use Kabsch's algorithm to get the orientation of the ball (the Kabsch algorithm computes the optimal rotation, w.r.t. the RMSE, between 2 sets of vectors).

\subsection{Dot Detection}
\begin{figure}
    \centering
    \includegraphics[width=0.4\textwidth]{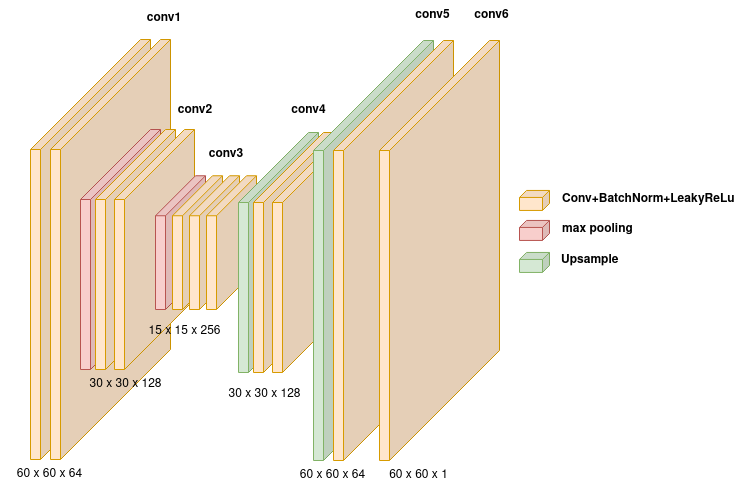}
    \caption{Dot detection CNN architecture}
    \label{fig:cnn_architecture}
\end{figure}

\begin{figure}
    \centering
    \includegraphics[width=0.4\textwidth]{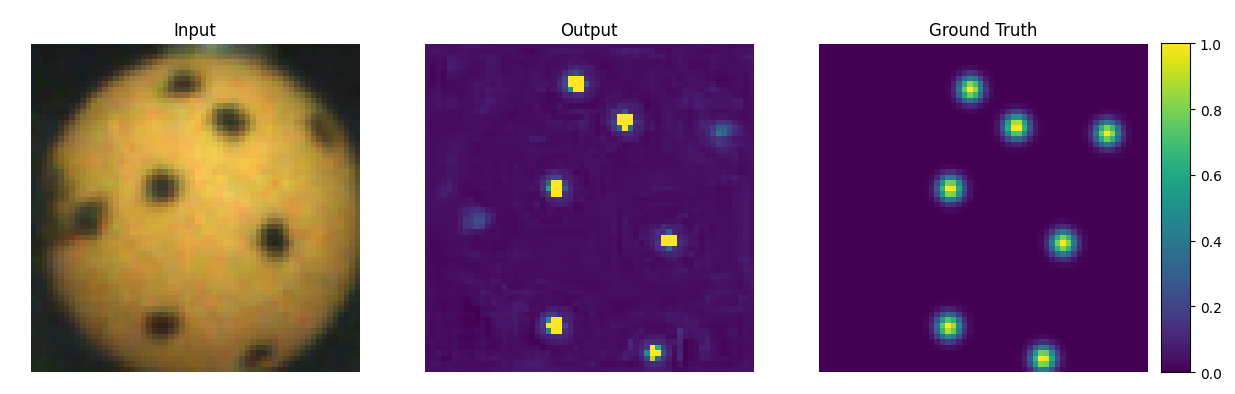}
    \caption{From left to right: Input Image, Output heatmap, Ground truth heatmap}
    \label{fig:cnn_elements}
\end{figure}

Conventional computer vision methods such as background subtraction, masking, blob detection were first tested to estimate the position of the dots on the image. They gave relatively good results after calibration but were also very sensitive to changes in the image (luminosity, contrast, color). It was thus abandoned for a CNN based on CenterNet \cite{duan2019a}. The exact architecture can be seen in Fig.\ref{fig:cnn_architecture}. This made the dot detection much more robust.
Using a CNN also enabled us to use balls with logo. Indeed, the CNN will learn to ignore the logo (as it can be seen in the leftmost images from Fig. \ref{fig:ball_reproj}), except in complex cases where the logo is only partially visible on the edge. This makes our method usable with any kind of ball, though the CNN would require some fine tuning for specific logos. In Fig.\ref{fig:cnn_elements}, we show an example of the input, output and ground truth of our CNN dot detector.

\subsection{Dataset}
\begin{figure}
    \centering
    \includegraphics[width=0.3\textwidth]{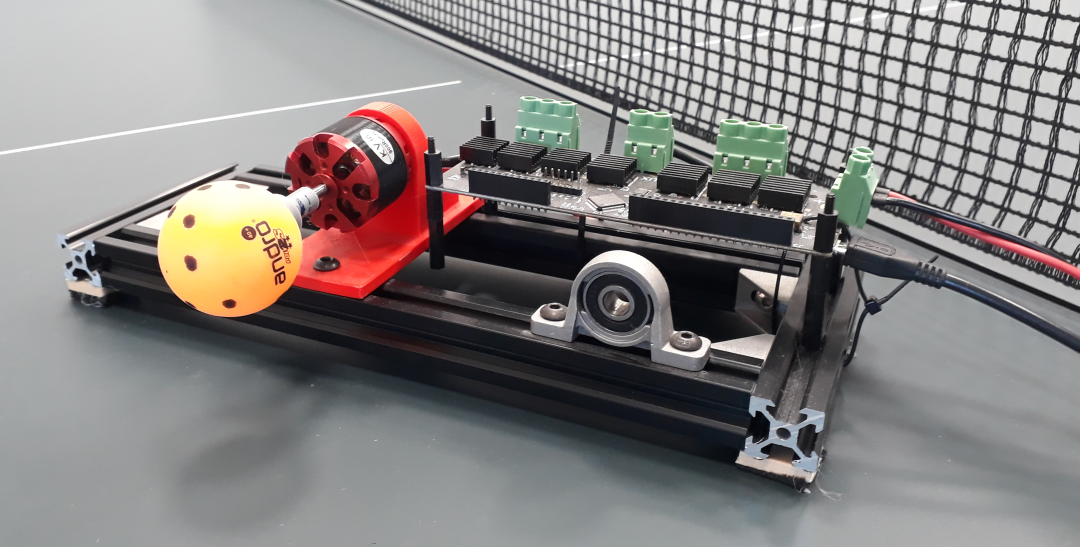}
    \caption{Ball spinner to generate training data}
    \label{fig:ball_spinner}
\end{figure}

In order to generate the dataset, the dotted ball was spun using a brushless DC motor (Fig.\ref{fig:ball_spinner}). The motor was controlled using a ODrive controller. The ODrive was chosen because of its Field Oriented Control (FOC) capabilities. With it, the spin's norm can be accurately measured (measurement noise of 0.1 rps) and used as ground truth.
To accurately get the rotation axis, the ball was installed so that the rotation axis corresponded with one of the drawn dots. Following this setup, we got ground truth (spin vector) and generated a spinning ball image dataset. These recordings were used as a benchmark tool for our spin estimation method and as a dataset to train the dot detection network. To get the dots' ground truth position on the image, the dots' reference 3D positions are rotated with the ball's calculated orientation. The ball's orientation is obtained by propagating an initial manually measured orientation with the corresponding spin. There is often some small offset between the estimated and actual dot positions. To compensate, the estimated dot position is corrected to the nearest local grayscale minimum which is a black dot. Additionally, the initial correction is reset every $\Delta t$ of measurements to avoid propagating error from the spin and initial orientation to the estimated ball orientations. A dataset of 80 000 samples was generated for balls possessing two different logos, with spin ranging from 0 to 150 rps. To make the dot detector more robust, the data was augmented by adding motion blur and changing brightness, contrast, saturation and hue.

\subsection{Bayesian Geometric Hashing}
Geometric hashing is an object recognition method which uses the spatial arrangement of keypoints to recognize an object. This method enabled us to identify the dots (get their index) even though they are all identical and have low resolution. 

Geometric hashing can be separated into two steps. First, a lookup table, the hash table, is generated from the reference object we want to recognize. Then this lookup table is used to identify the object.

\textbf{Generating the hash table:} the object's keypoints $\bm{D} =\{\bm{d_1}, \bm{d_2}, ...\}$ are transformed into a rotation, scale and translation invariant space called the hash space. This is achieved by using 2 keypoints (if working in 2D) as a basis to describe the relative position of the other features. This process is performed for every keypoint for every basis combination possible. In the end, we have a table that includes for each entry the keypoints used to form the basis and a vector $\bm{h}$ that is another keypoint position from the reference model transformed into the hash space using the specified basis.

\textbf{Recognition:} From the input image, the keypoints $\bm{D}$ are first extracted. From these keypoints, 2 are arbitrarily chosen as the basis to transform the other keypoints in the hash space. Each transformed keypoint $\bm{\phi}$ will give a vote to nearby hash values $\bm{h}$ in the hash space and thus a specific basis $\bm{basis_h}$. To do so, the hash values are binned and the transformed keypoint gives a vote to the hash value in the same bin as it. The basis with votes above a certain threshold will be preselected and their reprojection RMSE is computed. The basis with the smallest RMSE is chosen as the recognized model.

In our case, we are working with 3D points. We can calculate the 3D coordinates of the dots from their 2D image position since we know there are located on the ball's surface ($z = \sqrt{r - x^2 - y^2}$). The center of the ball and two dots are used as the basis needed to describe the other dots. 

The traditional geometric hashing method uses binning for the basis vote. However, a bayesian variant also exists \cite{wolfson1997}. Instead of preselecting the basis with the most votes, we choose the basis with the highest likelihood. This makes the geometric hashing much more robust because continuous probability distribution assures the continuity of the "voting", contrary to binning.

\begin{algorithm}
\caption{Geometric Hashing: Generating hash table}\label{alg:generating_hash_table}
\begin{algorithmic}
\Require  Reference dot positions: $\bm{D} = [\bm{d}_1, \bm{d}_2, ...] \in \mathbb{R}^{3 \times n}$
\State $\bm{hash\_table} = []$
\State $\bm{basis\_used} = []$
\For{$\bm{d}$ in $\bm{D}$}
    \For{$\bm{d'}$ in $\bm{D}\setminus\bm{d}$}
        \State $\bm{basis} = [\bm{d}, \bm{d'}, \bm{d} \times \bm{d'}]$
        \For{$\bm{d''}$ in $\bm{D} \setminus \{\bm{d}, \bm{d'}\}$}
            \State $\bm{h_i} = \bm{basis}^{-1} \cdot \bm{d''}$
            \State Append  $\bm{h_i}$ to $\bm{hash\_table}$
            \State Append $\bm{basis}$ to $\bm{bases\_used}$
        \EndFor
    \EndFor
\EndFor
\end{algorithmic}
\end{algorithm}

\begin{algorithm}
\caption{Bayesian Geometric Hashing: Recognition}\label{alg:bayesian_geometric_hashing}
\begin{algorithmic}
\Require Input dot position: $\bm{D} = [\bm{d}_1, \bm{d}_2, ...] \in \mathbb{R}^{3 \times n}$
\State $\bm{basis} = [\bm{d}_1, \bm{d}_2, \bm{d}_1 \times \bm{d}_2]$
\State $\bm{D'} \gets D \setminus \{\bm{d_1}, \bm{d_2}\}$
\State $\bm{\Phi} \gets \bm{basis}^{-1} \cdot \bm{D'}$ \Comment{Transform the dots into hash space}
\State $\bm{poss\_bases} = []$ \Comment{List of possible basis}
\For{$\bm{\phi} \in \bm{\Phi}$}
\State $\bm{H} \gets NearestHashValues(\bm{\phi})$
\For{$\bm{h}$ in $\bm{H}$}
\If{$\bm{basis_{h}}$ not in $\bm{poss\_bases}$}
\State Append $\bm{basis_{h}}$ to $\bm{poss\_bases}$
\EndIf
\State $\bm{score_{basis_{h}}} += p_{\bm{\phi}}(\bm{h}) $
\EndFor
\EndFor
\State $\bm{selected\_bases}$ are $\bm{basis}$ where $score \geq threshold$
\For{$\bm{basis}$ in $\bm{selected\_bases}$}
\State $\bm{rot} \gets Kabsch(\bm{basis}, \bm{D})$
\State $error \gets reprojection\_error(rot, \bm{D}$)
\EndFor
\State{Returned $\bm{rot}$ is the one with the least error}
\end{algorithmic}
\end{algorithm}

\begin{figure}
  \begin{subfigure}[b]{0.24\textwidth}
    \includegraphics[width=\textwidth]{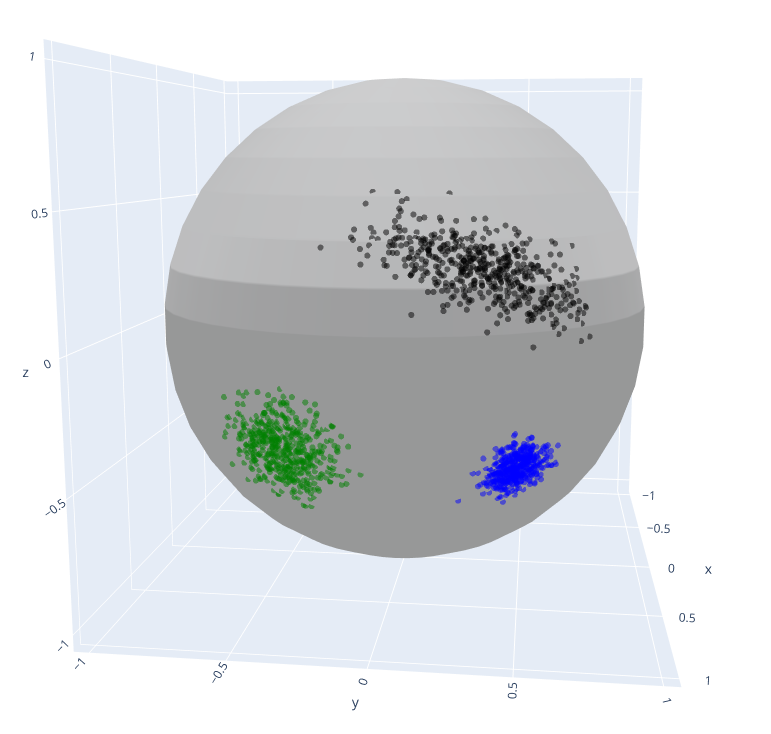}
    \caption{Sphere space}
    \label{fig:kent_dist}
  \end{subfigure}
  \begin{subfigure}[b]{0.24\textwidth}
    \includegraphics[width=\textwidth]{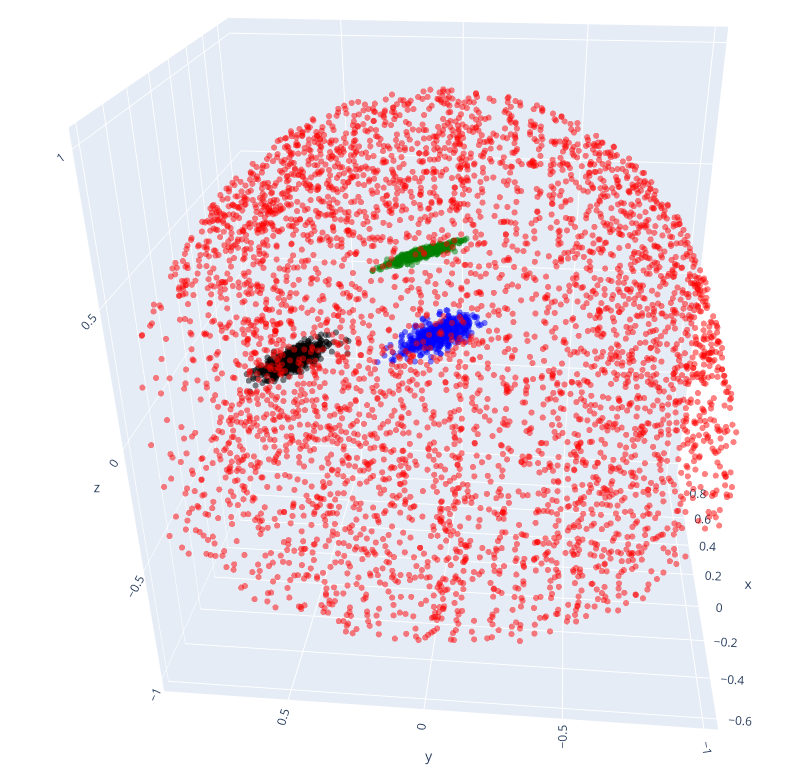}
    \caption{Hash space}
    \label{fig:geo_hashing}
  \end{subfigure}
  \caption{\textbf{Kent Distributions}: Samples are plotted on the sphere for different positions (Black: $\kappa=100$, $\beta=40$, Blue: $\kappa=300$, $\beta=0$, Green: $\kappa=100$, $\beta=0$)} The samples were transformed to the hash space using two other keypoints. The reds are all the hash values generated for the hash table.
\end{figure}

In the algorithm \ref{alg:bayesian_geometric_hashing}, $p_{\bm{\phi}}(\bm{h})$ represents the likelihood of the feature $\bm{\phi}$ (dot transformed into the hash space with the function $f(\bm{x}) = \bm{B}^{-1} \cdot \bm{x}$) corresponding to the hash value $\bm{h}$ (reference dot transformed into the hash space). We want this likelihood to represent the dot's position uncertainty in the hash space. To do so, we use the change of variable $f$ that will transform the dot position uncertainty on the sphere surface into the hash space equivalent, as shown in Eq.\ref{eq:variable_change}. The uncertainty can be due to multiple factors: dot misdrawn, CNN dot detector not accurate enough, motion blur. We use the Kent distribution to model the dot position uncertainty on the sphere $p_{\bm{d}}(\bm{x})$. The Kent distribution can be viewed as a normal distribution on a sphere's surface and is described as follows:

\begin{equation}
    k_{\bm{d}}(\bm{x})=\frac{1}{\mathrm{c}(\kappa, \beta)} \exp \left\{\kappa \gamma_{1}^{T} \cdot \bm{x}+\beta\left[\left(\gamma_{2}^{T} \cdot \bm{x}\right)^{2}-\left(\gamma_{3}^{T} \cdot \bm{x}\right)^{2}\right]\right\}
\end{equation}

where: \begin{equation}
        c(\kappa, \beta)=2 \pi \sum_{j=0}^{\infty} \frac{\Gamma\left(j+\frac{1}{2}\right)}{\Gamma(j+1)} \beta^{2 j}\left(\frac{1}{2} \kappa\right)^{-2 j-\frac{1}{2}} I_{2 j+\frac{1}{2}}(\kappa)
\end{equation} 

and $I_v(\kappa)$ is the modified Bessel function, $\Gamma (\cdot )$ is the gamma function, $\bm{\gamma_1}, \bm{\gamma_2}, \bm{\gamma_3}$ are orthogonal unit vectors representing the mean, major, and minor axes respectively of the pdf (they are calculated from $\bm{d}$), $\kappa$ determines the concentration of the pdf and $\beta$ determines the eccentricity of the pdf. These hyperparameters were empirically set to $\kappa=500$ and $\beta = 0$ so that the pdf corresponds to the dot's size and shape. Examples of Kent distributions with different parameters are shown in Fig.\ref{fig:kent_dist}.

To obtain $p_{\bm{\phi}} (\bm{h})$, we transform the Kent distribution in the hash space as follows:

\begin{equation}
\begin{split}
    p_{\bm{\phi}} (\bm{h}) &=  p_{\bm{d}}(f^{-1}(\bm{h})) |det \frac{\partial f^{-1}(\bm{h})}{\partial \bm{h}}|\\
    &=  p_{\bm{d}}(\bm{B}\cdot \bm{h}) |det(\bm{B})| 
\end{split}
\label{eq:variable_change}
\end{equation}

The norm of the determinant of the jacobian of the inverse function ensures that the volume of the probability distribution is maintained during the variable change.

\begin{figure}
    \centering
    \includegraphics[width=0.5\textwidth]{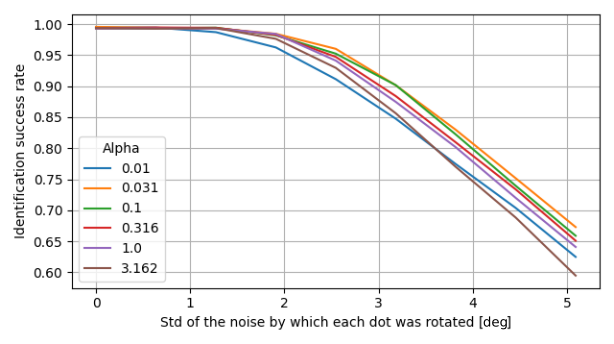}
    \caption{Geometric hashing sensitivity to inaccurate dot position}
    \label{fig:geo_hash_sensitivity}
\end{figure}

However, we need to add a "projection" likelihood $n(\bm{x})$ in order for the dot position uncertainty pdf $p_{\bm{d}}$ to work in the 3D hash space (the Kent distribution operates on the 2D sphere manifold). The projection $n(\bm{x})$ is set as the likelihood of a dot being on the surface of the sphere (norm of 1).

\begin{equation}
    p_{\bm{d}}(\bm{x}) = n(\bm{x}) k_{\bm{d}}(\bm{x})
    \label{eq:ll_feature}
\end{equation}

where:

\begin{equation}
n(\bm{x}) = \frac{1}{\alpha \sqrt{2 \pi}} \exp \left(-\frac{1}{2} \frac{(||\bm{x}||-1)^2}{\alpha^2}\right)
\end{equation}

and $\alpha$ is the standard deviation of our "projection" component. $\alpha = 0.03$ was chosen to maximize the identification rate of the geometric hashing and is visualized in Fig.\ref{fig:geo_hash_sensitivity}. We indeed investigated the sensitivity of the bayesian geometric hashing using a Monte Carlo test. The method used is explained in more detail in the next section.

\subsection{Pattern generation}
\begin{figure}
    \centering
    \includegraphics[width=0.4\textwidth]{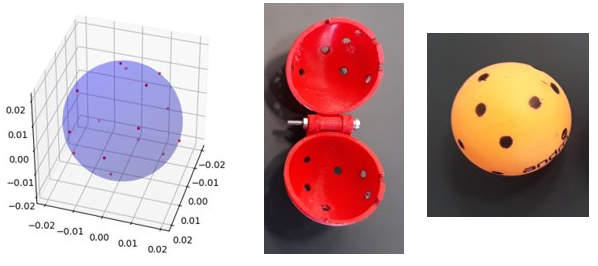}
    \caption{From left to right: Generated pattern, 3D printed mold, ball with dot pattern}
    \label{fig:3dprint}
\end{figure}

The dot pattern is drawn on the ball using a 3D printed stencil (Fig.\ref{fig:3dprint}). This enabled us to accurately and reliably position the dots on the ball's surface. This method also allows us to easily and cheaply make many balls for the robot table tennis setup.

The number of dots was arbitrarily set to 20. This gave an average of more than 3 points visible from any angle, which is necessary to use geometric hashing. The dots' position was on the other hand not arbitrary. With geometric hashing, objects are recognized thanks to the spatial configuration of their features. This means that all patterns are not equal and that some will have better robustness to noise. 

A pattern's robustness was evaluated by its success rate using Monte Carlo testing. A single test proceeds as follows: first, a random rotation is applied to the original dot configuration and only the visible dots ($d_z  > 0$ assuming that the $\vec{z}$ is pointing towards us) are selected. This is done to test out different viewing angles. Noise is then added to each visible dot individually by applying a small random rotation. The generated set of dots is finally passed to the geometric hashing. Patterns were evaluated on 10 000 samples with a noise of $\sigma = 3\degree$. This noise was thought to represent dot position error best. 

A gradient-based optimization approach was chosen to find the best pattern. We optimized the spherical coordinates $(\theta_i, \Phi_i)$ of dots located on the unit sphere by maximizing the mean nearest-neighbor distance of the features in the hash space. This would make the features in the hash space more distinguishable and thus more robust to noise. A dot pattern with a 98\% identification success rate was thus generated for a noise with a standard deviation of 3 \degree as can be seen in Fig.\ref{fig:geo_hash_sensitivity}.

\section{SPIN ESTIMATION} \label{spinestimation}

\subsection{Spin regression}
Though the dot detection and the bayesian geometric hashing have a high identification success rate, misidentification is not impossible. We couple RANSAC to the spin regression to make our algorithm robust to possible outliers. For the spin regression, we use QuateRA \cite{dealmeida2020a}, a quaternion-based spin regression algorithm. The rotation plan is calculated from the measurements using SVD. This gives us the rotation axis as the vector orthogonal to the rotation plane. Each quaternion is then projected onto the rotation plane. RANSAC is used here to select the valid measurements for the rotation plane calculation. The spin is then estimated with a linear regression from successive rotation angles. This approach is very similar to  \cite{tebbe2020}. The main difference is that the rotation plane is algebraically calculated using SVD in QuateRA, which is the least-square solution. In \cite{tebbe2020}, the rotation plane is the plane that intersects three logo positions. The three logo positions are chosen to minimize the projection error for the other logo positions. This makes our spin regression faster and more accurate than \cite{tebbe2020}.

\section{EXPERIMENT} \label{experiments}

\subsection{Setup}
The ball images are captured with a Grasshopper3 GS3-U3-23S6C camera at a framerate of 350 fps with a resolution of 1900x400. The exposure time was set to $250  \; \mu  s$. The ball's region of interest is isolated using the method described in \cite{tebbe2019}, and we then obtained images have a resolution of (60x60). Our spin estimation method, SpinDOE, is run on a computer equipped with a NVIDIA GeForce GTX 1080 Ti GPU and an Intel(R) Core(TM) i7-8700K CPU @ 3.70GHz.

\subsection{Orientation estimation}
\begin{figure}
    \centering
    \includegraphics[width=0.4\textwidth]{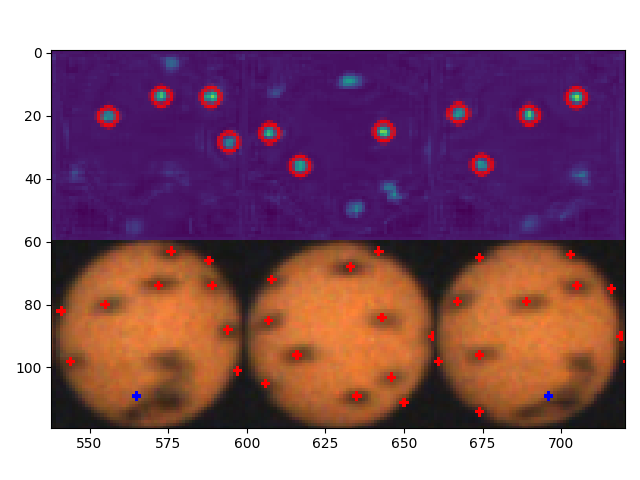}
    \caption{Ball spinner test case with 150 rps - \textbf{Row 1:} Generated heatmap, the dots located with the blob detection are circled in red; \textbf{Row 2:} input image on which we projected the estimated dots' position (red dots are estimated dot position and the blue dot is the estimated logo position)}
    \label{fig:ball_reproj}
\end{figure}

We first tested our orientation estimation method. We generated a benchmark of ball images and orientations thanks to the ball spinner. We give a visual example of the DOE output in Fig.\ref{fig:ball_reproj}. As shown, the CNN ignores the logo and outputs only high values in the heatmap for the dots. We can also notice that the CNN detects dots on the edge of the ball; this would not be the case using traditional computer vision algorithms.

\begin{figure}
    \centering
    \includegraphics[width=0.5\textwidth]{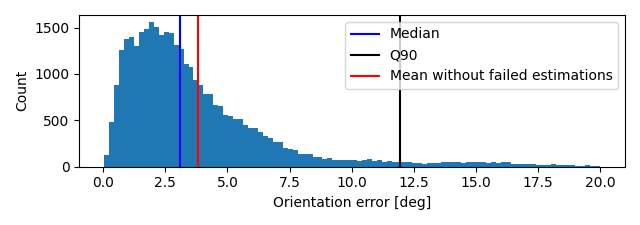}
    \caption{DOE's error distribution (range is limited between 0\degree and 20\degree for better readability )}
    \label{fig:orientation_benchmark}
\end{figure}

In Fig.\ref{fig:orientation_benchmark}, we plot the distribution of the orientation estimation error. As expected, there are cases where the orientation estimation fails. This is mostly due to the presence of the logo on the ball, which can sometimes be detected as one or two dots. The DOE is also sensitive to the ball being centered in the image. Indeed, the center of the image is assumed to be the center of the ball and the 3D position of the dots are calculated accordingly. There are also some estimation failures when only two or three dots are detected. Still, the failure rate (error above 20 \degree) is of 7\% on our whole benchmark dataset. This gives us a mean error of $2.3 \pm 2.3 $ \degree if the identification failures are ignored.

Regarding speed, the DOE takes $0.043 \pm 0.002$ s to process 10 images, the required number of ball images for the spin regression.

\subsection{Spin estimation}
We estimated the accuracy of our spin estimation method with our ball spinner benchmark. We restricted ourselves to using ten images as input representing a measurement duration of 26 ms.
\begin{figure}
    \centering
    \includegraphics[width=0.5\textwidth]{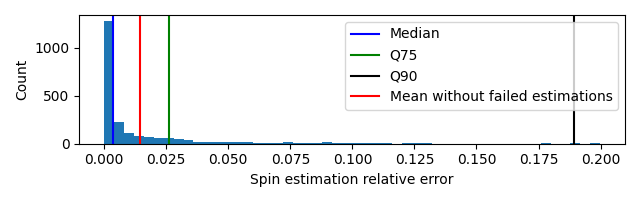}
    \caption{SpinDOE's relative error distribution (range is limited between 0 and 0.20 for better readability )}
    \label{fig:spin_benchmark}
\end{figure}
We observe in Fig.\ref{fig:spin_benchmark} that SpinDOE gives 90\% of the time a relative error lower than 0.20. Most erroneous spin estimations come from low spin values (below 10 rps) because of the increase of relative error and from high spins (above 140 rps) because the angle unwrapping in the spin regression fails.
The RANSAC spin regression takes $0.017 \pm 0.002\;s$ to run for a ten images input. In total, SpinDOE requires $0.061 \pm 0.003\;s$ to give out a spin value. This is enough to use SpinDOE in realtime.

\subsubsection{Constant spin assumption}
 One of the most common assumptions in robotics for table tennis is that the spin is constant. However, to our knowledge, this assumption has never been empirically proven. From a theoretical point of view, the air applies a viscous torque on all rotating spheres \cite{lei2006}, which slows down their spin. The ball's spin dynamic can be described by the following equation:
 
\begin{equation}
I \frac{\omega(t)}{d t} =  \frac{2}{4}m r^3\frac{ \omega(t)}{d t}= T_{viscous} = -8 \pi \nu r^3 \boldsymbol{\omega}(t)
\label{eq:viscous_torque}
\end{equation}
where $I$ is the ball's inertia, $\nu$ is the air viscosity, $m$ is the ball's mass and $r$ is the ball's radius.

Solving this ODE gives us the spin evolution:
\begin{equation}
 \begin{split}
      \omega(t)& =\omega(t_0) \exp \left(-\frac{12 \pi \nu r}{m} t\right)\\
      & \approx \omega(t_0) - \omega(t_0)\frac{12 \pi \nu r}{m}t 
\end{split}   
\end{equation}

with the dampening coefficient: $-\frac{12 \pi \nu r}{m} \approx-0.005$ (theoretical value using $\nu=1.81 \times10^{-5}$ kg/(m·s), $m=2.7g$ and $r=20$mm).

\begin{figure}
    \centering
    \includegraphics[width=0.45\textwidth]{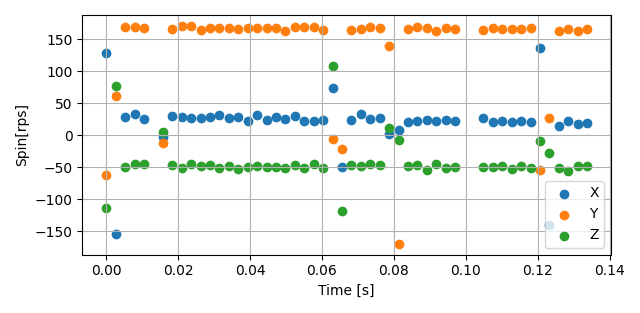}
    \includegraphics[width=0.45\textwidth]{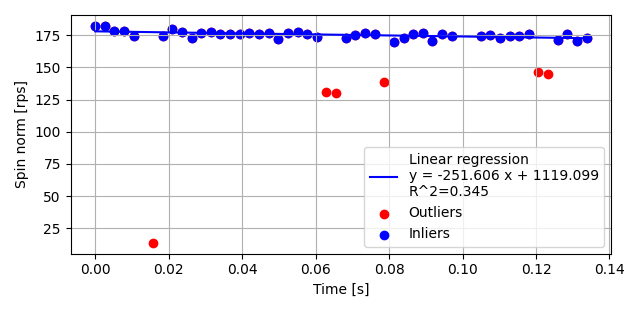}
    \caption{Spin calculated from the difference between 2 successive ball orientations}
    \label{fig:long_shot}
\end{figure}

In Fig.\ref{fig:long_shot}, we show the spin of a ball shot with a Butterfly Amicus ball throwing machine with the maximum spin and velocity setting. As we can see with the plot of the norm, the spin indeed decreases and a linear regression shows us that the theoretical value for the spin dampening is vastly underestimated. However, the spin decrease is so slight relative to the flight duration that the constant spin assumption holds.

We try to evaluate the spin dampening coefficient for several shots and obtain a mean value of $0.091\pm 0.03 $. This shows that the theoretical value is clearly underestimated.


\section{DATASET} \label{dataset}
Using SpinDOE, we generated a dataset of table tennis ball trajectories. The positions are recorded at 145 Hz and the spin is recorded only once, making it only valid until the first bounce. To our knowledge, no other similar dataset is publicly available. The position of the ball is captured using the method described in \cite{tebbe2019}. Two cameras are used for triangulating the ball's position. The ball's position in the image is extracted using standard computer vision algorithms (background subtraction, mask generation and blob detection).

\begin{figure}
    \centering
    \includegraphics[width=0.35\textwidth]{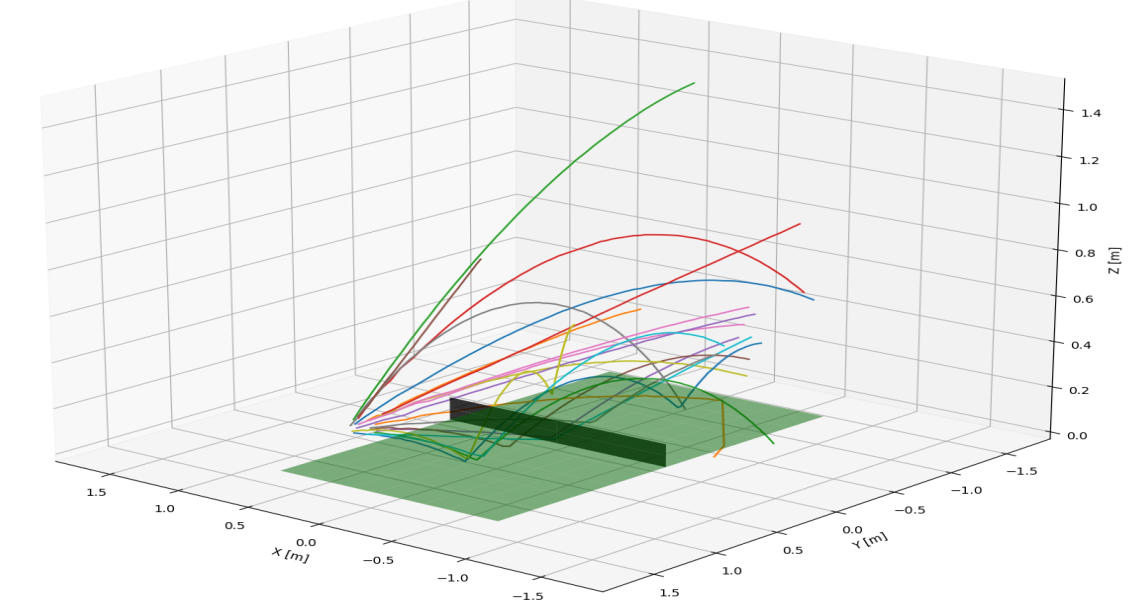}
    \caption{Subsample of the trajectories from the recorded dataset}
    \label{fig:dataset_trajectories}
\end{figure}

The dataset contains 200 trajectories, a subsample of which can be seen in Fig. \ref{fig:dataset_trajectories}. The maximum velocity and spin observed were respectively 11 m/s and 150 rps. The balls were shot with the Amicus ball thrower for which the shooting settings where uniformly sampled.

This dataset can have multiple applications: benchmarking trajectory prediction algorithms, checking that algorithms that estimate spin from ball trajectories indeed work or building a bayesian table tennis simulator for a smaller sim2real gap in reinforcement learning, for example.

\section{CONCLUSIONS}
Though our method can not be used for measuring spin during official table tennis matches, it can be very useful for research. We provide a dataset of ball trajectories with spin for that purpose. SpinDOE does not require high resolution or high fps compared to other methods. It is also robust to motion blur and does no suffer from hidden markers, as logo-based methods do. It is most of all very accurate, e.g. one can observe the effect of the viscous torque slowing down the ball's spin.
This method is used in the context of a table tennis robot, but it could also be applied to other ball sports such as tennis, handball or football.


\bibliographystyle{IEEEtran}
\bibliography{IEEEabrv,tt_spin}

\end{document}